\documentclass[10pt,twocolumn,letterpaper]{article}

\usepackage{cvpr}
\usepackage{times}
\usepackage{epsfig}
\usepackage{graphicx}
\usepackage{amsmath}
\usepackage{amssymb}
\usepackage{algorithm,algorithmic}
\usepackage{helvet}
\usepackage{courier}
\usepackage{url}
\usepackage{latexsym}
\usepackage{booktabs} 
\usepackage{color}
\usepackage{multirow}
\usepackage{bbm}
\usepackage{float}
\usepackage{caption}
\usepackage{subfig}
\usepackage{color}
\usepackage{wrapfig}
\usepackage[T1]{fontenc}
\usepackage[utf8]{inputenc}
\usepackage{amsfonts}

\usepackage[pagebackref=true,breaklinks=true,letterpaper=true,colorlinks,bookmarks=false]{hyperref}

\newcommand{\Amat}[0]{{{\bf A}}}
\newcommand{\Bmat}{{\bf B}}
\newcommand{\Cmat}{{\bf C}}
\newcommand{\Dmat}{{\bf D}}

\newcommand{\Fmat}[0]{{{\bf F}}\xspace}

\newcommand{\Imat}{{\bf I}}

\newcommand{\Umat}[0]{{{\bf U}}}
\newcommand{\Vmat}[0]{{{\bf V}}}

\newcommand{\av}[0]{{\boldsymbol{a}}}
\newcommand{\bv}[0]{{\boldsymbol{b}}}
\newcommand{\cv}[0]{{\boldsymbol{c}}}

\newcommand{\ev}[0]{{\boldsymbol{e}}\xspace}

\newcommand{\uv}{\boldsymbol{u}}
\newcommand{\vv}{\boldsymbol{v}}

\newcommand{\xv}{\boldsymbol{x}}
\newcommand{\yv}{\boldsymbol{y}}
\newcommand{\zv}{\boldsymbol{z}}

\newcommand{\Phimat}{\boldsymbol{\Phi}}

\newcommand{\lambdav}[0]{{\boldsymbol{\lambda}}}
\newcommand{\muv}[0]{{\boldsymbol{\mu}}}

\newcommand{\Lcal}{\mathcal{L}}
\newcommand{\Ncal}{\mathcal{N}}

\newcommand{\Tcal}{\mathcal{T}}

\newcommand{\Ccal}{\mathcal{C}}

\cvprfinalcopy 

\def\cvprPaperID{3311} 

\ifcvprfinal\fi
\begin{document}

\title{Nonlocal Low-Rank Tensor Factor Analysis for Image Restoration}

\author{
Xinyuan Zhang$^\dagger$\\
Duke University\\
Durham, NC, USA\\
{\tt\small xy.zhang@duke.edu}
\and
Xin Yuan$^*$\\
Nokia Bell Labs\\
Murray Hill, NJ, USA\\
{\tt\small xyuan@bell-labs.com}
\and
Lawrence Carin\\
Duke University\\
Durham, NC, USA\\
{\tt\small  lcarin@duke.edu}
}

\maketitle

\begin{abstract}
Low-rank signal modeling has been widely leveraged to capture non-local correlation in image processing applications. We propose a new method that employs low-rank tensor factor analysis for tensors generated by grouped image patches. The low-rank tensors are fed into the alternative direction multiplier method (ADMM) to further improve image reconstruction. The motivating application is compressive sensing (CS), and a deep convolutional architecture is adopted to approximate the expensive matrix inversion in CS applications. An iterative algorithm based on this low-rank tensor factorization strategy, called NLR-TFA, is presented in detail. Experimental results on noiseless and noisy CS measurements demonstrate the superiority of the proposed approach, especially at low CS sampling rates.
\let\thefootnote\relax\footnote{{{$^\dagger$This work was performed when Xinyuan Zhang was a summer intern at Nokia Bell Labs in 2017.}}}
\let\thefootnote\relax\footnote{{$^*$Corresponding author.}}
\end{abstract}

\section{Introduction}
Inspired by the nonlocal self-similarity of image patches~\cite{Buades05NLM}, various image processing algorithms have been proposed to investigate the 
{\em low-rank} property of image patch groups~\cite{Dabov07BM3D,Dong14_NLRCS,Gu14WNNM,Nie12Spnorm,Xie16WSPnorm}.
In general, these methods first select a reference patch and then search for similar patches across the image to form a group.
Following this, the patches in this group are {\em vectorized} and stacked to a matrix. Since these patches are similar, the constructed matrix has a low-rank property.
Via performing this low-rank model on every (overlapping) patch in the image, state-of-the-art image restoration results have been achieved.
 
One common issue in the above algorithms is that the original two-dimensional (2D) patches are vectorized to construct the group matrix, which loses the spatial structure within the image patch. We propose a tensor based algorithm to retain this structure while still leveraging the advantages of low-rank patch models.
Tensor factorization methods~\cite{Kolda09Tensor} offer a useful way to learn latent structure from complex multiway data, and have been used in image processing tasks~\cite{Xie_2016_CVPR}. 
These methods decompose the tensor data into a set of factor matrices (one for each mode or “way” of the tensor), that can be used as a latent feature representation for the objects in each of the tensor modes~\cite{Rai14ICML}. Recently tensor approaches have been applied in computer vision, such as image denoising~\cite{peng2014decomposable}, and video denoising~\cite{wen2017joint}.
We consider the general image restoration problem with specific applications to compressive sensing~\cite{Donoho06CS} using the tensor factor analysis (TFA) approach.
In particular, instead of vectorizing the image patches, we impose {\em low-rank} TFA to the 3D image patch groups.

\begin{figure}[t!]
	\centering
	\includegraphics[scale = 0.3]{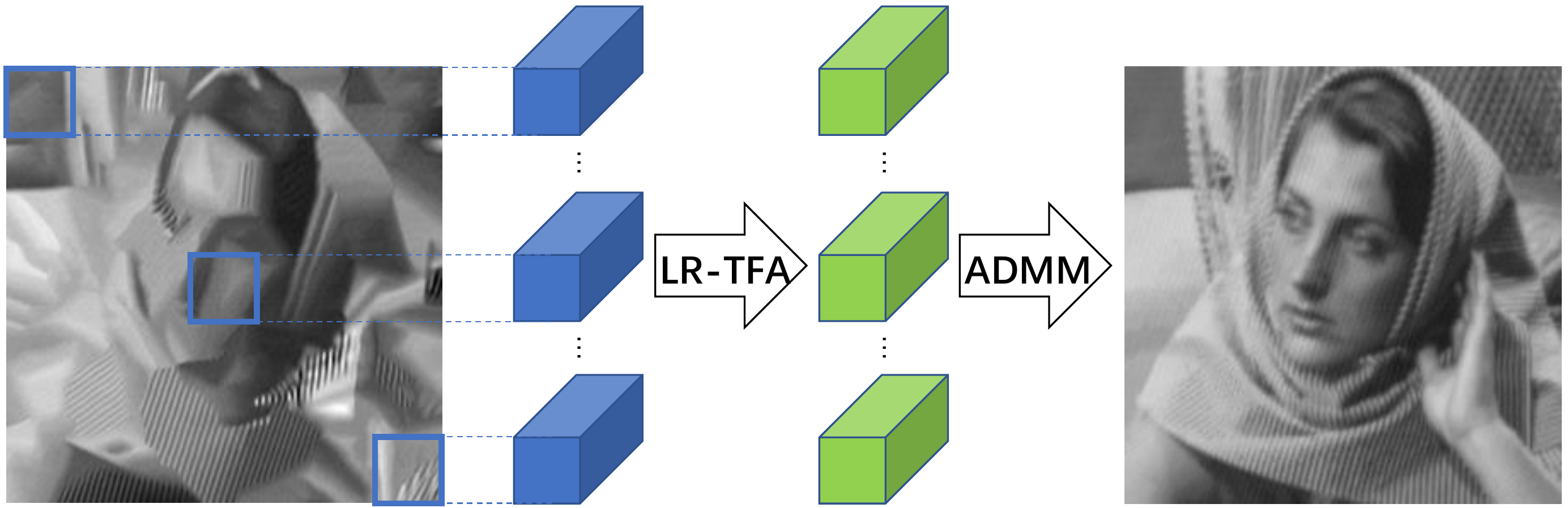}
	
	\caption{An illustration of the NLR-TFA model. The left image is the initial estimate using some fast algorithm. Each tensor is concatenated by a reference image patch and its similar patches. Blue cubes represent full-rank tensors; green cubes represent low-rank tensors. The right image (PSNR=$26.15$dB) is reconstructed using NLR-TFA from compressive sensing (CS) measurements at a CS rate of only $0.02$. Notice how the proposed method can restore rich semantic content and fine structure of the image, even when the sensing rate is extremely low.
	}
	\label{framework}
\end{figure}

To be concrete,
the image restoration problem aims to estimate the latent clean image $\xv$ given the observation $\yv$ (the corrupted/compressive measurement) and the measurement matrix $\Phimat$, which can be formulated as 
\begin{equation}
\yv = \Phimat\xv+\ev,
\end{equation}
where $\ev$ denotes the measurement noise and is usually modeled by $\ev\sim\Ncal(0,\sigma^2\Imat)$. 
The problem has different variants for different $\Phimat$: when $\Phimat$ is the identity matrix, this is a denoising problem~\cite{Elad06_denoising}; when $\Phimat$ is a diagonal matrix whose diagonal elements are either 1 or 0, keeping or removing corresponding
pixels, this becomes image inpainting~\cite{Zhou09NIPS}; and when $\Phimat\in\mathbb{C}^{M\times N}$ and $M\ll N$, this is a compressive sensing (CS) problem.

We focus on the CS problem~\cite{Duarte08SPM,Lustig08SPM}, and consider the image reconstruction with a very limited number of compressive measurements~\cite{Yuan17_blockCS_ICIP,Yuan15Lensless,Yuan16SJ,Yuan18OE}. A framework of the proposed NLR-TFA method is shown in Fig.~\ref{framework}.
There has been recent interest in using deep learning technique for CS problems \cite{mousavi2015deep}. A reconstruction network is developed in~\cite{Kulkarni_2016_CVPR,Yuan18OE}, and an ADMM-net is proposed in~\cite{ADMM_NIPS2016}; while the former paper focused on block-wise CS, the latter one only considered CS-MRI~\cite{Lustig08SPM}.  
By contrast, we develop a tensor-based framework for general CS applications. Further, to overcome the large-scale matrix inversion in CS, we further propose to pre-train a deep convolutional neural network to approximate the expensive matrix inversion operator~\cite{wei2017inner}.

The remainder of this paper is organized as follows.
Section~\ref{Sec:BG} presents background knowledge on tensors and CS. Section~\ref{Sec:Algo} derives our algorithm. Extensive results are provided in Section~\ref{Sec:Results} to demonstrate the superiority of the proposed algorithm relative to other leading approaches, and Section~\ref{Sec:Con} concludes the paper.

\section{Background \label{Sec:BG}}

\subsection{Tensor Decomposition}
Tensors (of order higher than two) are arrays indexed by three or more indices. Specifically, matrices are two-way arrays, and tensors are three- or higher-way arrays. In the sequel, we mainly focus on $3\text{-}$way tensors and everything naturally generalizes to higher-order tensors.

A $rank\text{-}1$ $3\text{-}$way tensor $\Tcal\in\mathbb{R}^{I\times J\times K}$ is an outer product of $3$ unit vectors $\av\in\mathbb{R}^I$, $\bv\in\mathbb{R}^J$, and $\cv\in\mathbb{R}^K$:
\begin{align}
\Tcal &= \lambda \av\otimes\bv\otimes\cv\\
\Tcal(i,j,k)&=\lambda \av(i)\bv(j)\cv(k)\nonumber
\end{align} 
where $\otimes$ is the vector outer product and $\lambda$ is a scalar.

The $rank$ of a tensor $\Tcal$ is the minimum number of $rank\text{-}1$ tensors that sum to $\Tcal$. Therefore, a $rank\text{-}R$ $3\text{-}$way tensor $\Tcal$ can be written as
\begin{align}\label{eq:LR-TFA}
\Tcal&=\sum_{r=1}^{R}\lambda_r\av_r\otimes\bv_r\otimes\cv_r\\
\Tcal(i,j,k)&=\sum_{r=1}^{R}\lambda_r \av_r(i)\bv_r(j)\cv_r(k)\nonumber
\end{align}
where unit vectors $\av_r$, $\bv_r$, and $\cv_r$ are tensor factors and $\lambda_r$ is a scalar that evaluates the significance of tensor factors $\av_r,\bv_r,\cv_r$. For brevity, $\Tcal=\sum_{r=1}^{R}\lambda_r\av_r\otimes\bv_r\otimes\cv_r$ can be denoted as $\Tcal=\langle\lambdav,\Amat,\Bmat,\Cmat\rangle$, where $\lambdav:=[\lambda_1,...,\lambda_R]$, $\Amat:=[\av_1,...,\av_R]$, $\Bmat:=[\bv_1,...,\bv_R]$, and $\Cmat:=[\cv_1,...,\cv_R]$. When the rank $R$ is low, the number of elements needed to represent $\Tcal$ is dramatically decreased after tensor decomposition, \ie, from $I\times K\times J$ to $(I+K+J)\times R$.

Practically, most tensor decomposition problems are NP-hard. However, in most real applications, as long as the tensor does not have too many components, and the components are not adversarially chosen, tensor decomposition can be computed in polynomial time \cite{sidiropoulos2017tensor}. The tensor-decomposition problem seeks to estimate $\Amat$, $\Bmat$, $\Cmat$ and coefficients $\lambdav$ from a tensor $\Tcal$. Adopting a least-squares fitness criterion, the problem is 
\begin{align}
\min_{\Amat,\Bmat,\Cmat,\lambdav}\|\Tcal-\sum_{r=1}^{R}\lambda_r\av_r\otimes\bv_r\otimes\cv_r\|_F^2
\end{align}
In this work, we employ Jenrich's algorithm \cite{leurgans1993decomposition} to solve this problem. In Algorithm \ref{Jenrich}, ``$+$'' denotes pseudo-inverse of a matrix.
\begin{center}
	\begin{algorithm}[htbp]
		\caption{Jenrich's Algorithm}
		\begin{algorithmic}[1]
			\REQUIRE Tensor $\Tcal$.
			\STATE   Pick two random unit vectors $\uv,\vv$.
			\STATE Compute $\mathcal{T}_{\uv}=\sum_{i=1}^{k}u_i\mathcal{T}[:,:,i]$;
			\STATE Compute $\mathcal{T}_{\vv}=\sum_{i=1}^{k}v_i\mathcal{T}[:,:,i]$;
			\STATE $\av_r$'s are eigenvectors of $\mathcal{T}_{\uv}(\mathcal{T}_{\vv})^+$, $\bv_r$'s are eigenvectors of $(\mathcal{T}_{\uv}^\top)(\mathcal{T}_{\vv}^\top)^+$;
			\STATE Given $\Amat$ and $\Bmat$, we can get $\cv_r$'s and $\lambda_r$'s by solving a linear system followed by normalization;
			\STATE $\textbf{Output:}$ Tensor factors $\av_r$'s, $\bv_r$'s, $\cv_r$'s and coefficients $\lambda_r$'s
		\end{algorithmic}
		\label{Jenrich}
	\end{algorithm}
\end{center}
\subsection{Compressive Sensing}
Compressive sensing is a signal acquisition technique that enables sampling a signal at sub-Nyquist rates~\cite{Donoho06CS}. In CS, a reconstruction algorithm is used to recover the original high-dimensional signal from a small number of random linear measurements. Taking compressive measurements can be viewed as a process of linearly mapping an $N$-dimensional signal vector $\xv$ to an $M$-dimensional measurement vector $\yv$, $M\ll N$, using a measurement matrix $\Phimat\in\mathbb{C}^{M\times N}$, \ie, $\yv=\Phimat\xv$. Since the matrix $\Phimat$ is rank-deficient, there exists more than one $\xv$ that yields the same measurement $\yv$. In this paper, the compressive sensing rate (CSr) is defined as CSr$=M/N$. 

To recover $\xv$, one searches for the vector that possesses a certain structure among all the vectors $\xv$ that satisfy $\yv\approx\Phimat\xv$. In the case of a sparse $\xv$, a popular method is to solve the optimization problem
\begin{equation}\label{eq:cs}
\xv=\arg\min_x\|\xv\|_1,\ s.t.\ \|\yv-\Phimat\xv\|\leq\epsilon.
\end{equation}
This problem is convex and known formally as basis pursuit denoising (BPDN). It has been shown in \cite{candes2005decoding}\cite{donoho2006compressed} that if $\xv$ is sufficiently sparse and $\Phimat$ satisfies certain properties, then the $s$-sparse signal can be accurately recovered from $m=\mathcal{O}(s\log(n/s))$ random linear measurements \cite{candes2006robust}. Equation (\ref{eq:cs}) can be equivalently translated to the following unconstrained optimization problem
\begin{equation}
\xv=\arg\min_x\|\yv-\Phimat\xv\|_2^2+\lambda\|\xv\|_1,
\end{equation}
where $\lambda$ is a regularization parameter. Various methods can be used to solve the above minimization problem~\cite{Yuan16ICIP_GAP,Yuan14TSP,Yuan17_CSCFA_ICIP}, and in this work we adopt the alternative direction multiplier method (ADMM) framework \cite{boyd2011distributed,lin2010augmented}.
Specifically, we consider the application of image CS~\cite{Duarte08SPM,Yuan15Lensless,Yuan16SJ}, and beyond sparsity, we propose a new tensor factorization approach to exploit the high-order structure in the image patches, seeking high reconstruction performance at extremely low CSr, \eg, CSr<0.05. Refer to Fig.~\ref{fig:foreman} for one example of reconstructed image using our proposed algorithm, compared with other leading algorithms at CSr$=0.02$ (with image size $256\times 256$).

\section{Method \label{Sec:Algo}}
We propose a new model that recovers compressively sensed images using low-rank tensor factor analysis and ADMM. First, we generate a tensor from the estimated image based on patch grouping. Then we impose low rankness on the tensor after tensor decomposition. This new low-rank tensor is fed into a global objective function which is solved by ADMM. 
These two steps are iteratively performed until satisfying some criterion. 
The complete algorithm is shown in Algorithm \ref{alg}.

\subsection{Patch-Based Low-Rank Tensor Factorization}
Given the observation $\yv$, we first obtain the estimated image $\hat{\xv}$ using some fast algorithm, \eg, the DCT or wavelet based algorithm~\cite{He09SPT,Yuan14TSP}. In denoising, $\hat{\xv}$ can be simply set equal to $\yv$. Then the estimated image is divided into $P$ overlapping patches $\{\xv_1,...,\xv_P\}$. The basic assumption underlying the proposed approach is that a sufficient number of similar patches can be found for any reference patch, a.k.a, the nonlocal self-similarity (NSS) prior~\cite{Buades05NLM}. For each reference patch $\xv_p$, we perform a $k$-nearest-neighbor search for the nonlocal similar patches across the image to form a group $\{\xv_{p,1},\xv_{p,2},...,\xv_{p,k}\}$, where $k$ is the number of similar patches (including the reference patch itself). Here, the Euclidean distance of pixel intensity is used as the metric to group patches. By concatenating the grouped patches on {\em the third dimension} in ascending order of Euclidean distances, we generate a tensor $\Tcal_p$ for reference patch $\xv_p$
\begin{align}
\Tcal_p = [\xv_{p,1},\xv_{p,2},...,\xv_{p,k}].
\end{align}
As $\xv_p$ has zeros distance to itself, it is always found as the leading patch in $\Tcal_p$, \ie, $\Tcal_p(:,:,1)=\xv_p$. Eventually we have $P$ tensors and each tensor corresponds to a reference patch. The coordinates of the grouped patches are also recorded for later image aggregation. Suppose the size of each patch is $m\times n$, then the size of generated tensor is $\Tcal_p \in {\mathbb R}^{m\times n\times k}$.
It has been shown that the grouped patches can be denoised by low-rank approximation~\cite{buades2016patch}. In this work, the low-rankness is imposed on the tensor by taking the most significant tensor factors after tensor decomposition
\begin{align}\label{eq:LR}
\Lcal_p=\sum_{r\in S_\ell}\lambda_r\av_r\otimes\bv_r\otimes\cv_r,
\end{align}
where $S_\ell$ selects the most significant $\ell$ tensor factors of $\Tcal_p$.
Therefore, $\Lcal_p$ has rank $\ell$. The significance of tensor factors is evaluated by $\lambda_r$'s in (\ref{eq:LR-TFA}).
\begin{figure}[t!]
	\centering
	\includegraphics[scale = 0.29]{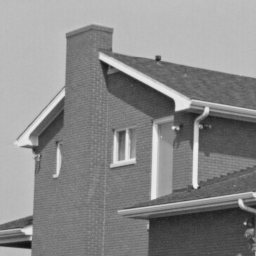}
	\includegraphics[scale = 0.29]{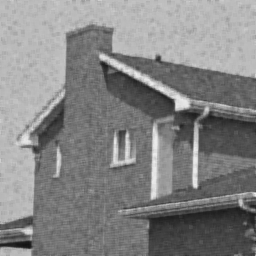}
	\includegraphics[scale = 0.29]{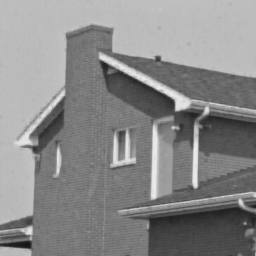}\\
	\vspace{0.03in}
	\includegraphics[scale = 0.29]{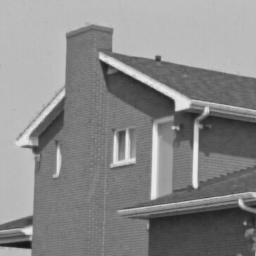}
	\includegraphics[scale = 0.29]{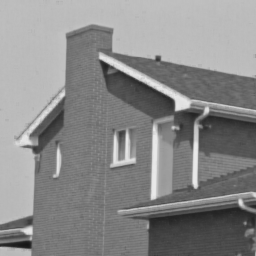}
	\includegraphics[scale = 0.29]{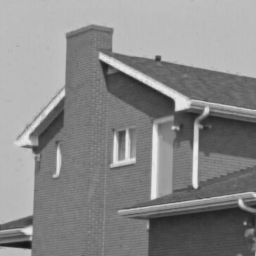}\\
	\vspace{0.03in}
	\includegraphics[scale = 0.29]{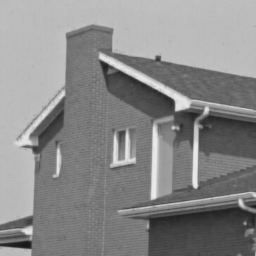}
	\includegraphics[scale = 0.29]{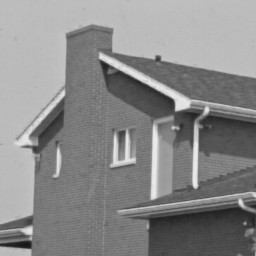}
	\includegraphics[scale = 0.29]{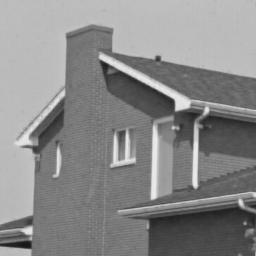}
	\caption{Aggregating low-rank tensors to images. The original image is on top left. The PSNR of images reconstructed from $rank$-$1$ to $rank$-$8$ tensors are $27.88$dB, $34.78$dB, $36.54$dB, $36.65$dB, $36.67$dB, $36.76$dB, $37.00$dB, and $37.09$dB respectively.}
	\label{low-rank}
\end{figure}

Under the assumption that the grouped patches have similar structure, $\Tcal_p$ has a low-rank property, which ensures that $\Tcal_p$ can be represented by a relatively low-rank tensor $\Lcal_p$. 
This low-rank imposition shares the same spirit with the hard thresholding algorithm~\cite{BLUMENSATH2009265}.
In Fig.~\ref{low-rank}, we impose low rankness on tensors with size $8\times 8\times 20$ generated from a clean image using the above approach, and then aggregate the tensors back to images. As can be seen, reconstructed images with low-rank tensors can accurately approximate the original image. The images aggregated from $rank$-$3$ to $rank$-$8$ tensors are highly similar (with PSNR merely increased by $0.55$ dB), which indicates that in this case $rank$-$3$ tensors are adequate to represent the original image.

\subsection{Image Recovery via ADMM}
With the reconstructed low-rank tensors $\Lcal_p$, the following optimization problem is proposed for CS recovery:
\begin{equation}\label{eq:obj}
{\hat \xv}=\arg\min_{\xv}\|\yv-\Phimat\xv\|_2^2+\eta\sum_p\|\tilde{\Tcal}_p\xv-\Lcal_p\|_F^2,
\end{equation}
where $\tilde{\Tcal}_p\xv=\Tcal_p$ denotes the tensor formed for each reference patch $\xv_p$ and $\eta$ is a regularization parameter. The closed-form solution for this quadratic optimization problem is
\begin{equation}\label{eq:x_close}
\xv=(\Phimat^H\Phimat+\eta\sum_p\tilde{\Tcal}_p^\top\tilde{\Tcal}_p)^{-1}(\Phimat^H\yv+\eta\sum_p\tilde{\Tcal}_p^\top\Lcal_p),
\end{equation}
where $H$ is the Hermitian transpose operation, $\sum_p\tilde{\Tcal}_p^\top\Lcal_p$ denotes the results of averaging all of the similar patches for each reference patch, and each entry of  $\sum_p\tilde{\Tcal}_p^\top\tilde{\Tcal}_p$ corresponds to an image pixel location whose value is the number of overlapping patches that cover this pixel location. In (\ref{eq:x_close}), the matrix to be inverted can be large, for which conjugate gradient descent (CG) is usually applied~\cite{chang2017one}. We adopt ADMM to solve this problem, introducing auxiliary variable $\zv$. Applying ADMM to (\ref{eq:obj}), we obtain the global objective function
\begin{align}\label{eq:admm}
({\hat \xv},{\hat \zv}) = \arg\min_{\xv,\zv}&\|\yv-\Phimat\xv\|_2^2+\beta\|\xv-\zv+\frac{\muv}{2\beta}\|_2^2\nonumber\\&+\eta\sum_p\|\tilde{\Tcal}_p\zv-\Lcal_p\|_F^2,
\end{align}
where $\muv$ is the Lagrange multiplier, and $\beta>0$ is the penalty parameter. Instead of minimizing $\xv$ and $\zv$ simultaneously, ADMM decomposes the problem into two subproblems that minimizes w.r.t $\xv$ and $\zv$, respectively. More specifically, the optimization problem in (\ref{eq:admm}) consists of the following iterations:
\begin{align}
\zv^{t+1}&=\arg\min_{\zv}\beta\|\xv^{t}-\zv+\frac{\muv^{t}}{2\beta}\|_2^2+\eta\sum_p\|\tilde{\Tcal}_p\zv-\Lcal_i\|_F^2,\\
\xv^{t+1}&=\arg\min_{\xv}\|\yv-\Phimat\xv\|_2^2+\beta\|\xv-\zv^{t+1}+\frac{\muv^{t}}{2\beta}\|_2^2,\\
\muv^{t+1}&=\muv^{t}+2\beta(\xv^{t+1}-\zv^{t+1}).\label{eq:mu}
\end{align}
Both $\xv$ and $\zv$ admit closed-form solutions. For fixed $\xv^{t}$ and $\muv^{t}$,
\begin{align}\label{eq:update_z}
\zv^{t+1}=(\eta\sum_p\tilde{\Tcal}_p^\top\tilde{\Tcal}_p+\beta\Imat)^{-1}(\beta\xv^{t}+\frac{\muv^{t}}{2}+\eta\sum_{p}\tilde{\Tcal}_p^\top\Lcal_p).
\end{align}
Then we use the updated $\zv^{t+1}$ to update $\xv$,
\begin{align}\label{eq:update_x}
\xv^{t+1}=\Umat^{-1}(\Phimat^H\yv+\beta\zv^{t+1}-\frac{\muv^{t}}{2}),
\end{align}
where $\Umat^{-1}=(\Phimat^H\Phimat+\beta\Imat)^{-1}$.

\begin{center}
	\begin{algorithm}[htbp]
		\caption{CS via Low-Rank TFA }
		\begin{algorithmic}[1]
			\REQUIRE The observation measurement $\yv$.
			
			\textbf{Initialization}:
			
			\STATE   Set parameters $\eta$, $\beta$, $\ell$, $m$, $n$, $k$, $I$, and $J$.
			
			\STATE		Pretrain $\Ccal_\theta$ for inverting matrix $\Phimat^H\Phimat+\beta\Imat$ using (\ref{eq:train});
				
			\STATE	Obtain an estimated image $\hat{\xv}$ from observation $\yv$ using a fast CS method;
				
			\STATE	Divide $\hat{\xv}$ into a set of overlapping patches;
				
			\STATE	Initialize $\xv^{(0)}=\hat{\xv}$;
			
			\textbf{Image Restoration}:
			\FOR{$i=0,1,...,I-1$}

			\STATE Divide $\xv^{(i)}$ into a set of overlapping patches with size $m\times n$;
			
			\STATE Form a set of tensors with size $m\times n\times k$ using patch block matching;
			\STATE Decompose tensors using Jenrich's Algorithm;
			\STATE Impose low rankness on tensors via (\ref{eq:LR}) and generate a set of $rank$-$\ell$ tensors;
			
			\textbf{ADMM}:
			
			\STATE Initialize $\muv^{0}=0$, $\xv^{0}=\xv^{(i)}$
			\FOR{$j=0,1,...,J-1$}
			
			\STATE Update $\zv^{j+1}$ via (\ref{eq:update_z});
			\IF{$\Phimat$ is a partial Fourier transform matrix}
			\STATE Update $\xv^{j+1}$ via (\ref{eq:fft2});
			\ELSE 
			\STATE Update $\xv^{j+1}$ via (\ref{eq:deep});
			\ENDIF
			\STATE 	Update $\muv^{j+1}$ via (\ref{eq:mu});
			\ENDFOR
			\STATE Update $\xv^{(i+1)}=\xv^{J}$
			\ENDFOR
		
			\STATE $\textbf{Output:}$ The reconstructed image $\hat{\xv}=\xv^{(I)}$.
		\end{algorithmic}
		\label{alg}
	\end{algorithm}
\end{center}

\subsubsection{Pretrained Deep Convolutional Architectures}
In (\ref{eq:update_x}), the term $\Umat^{-1}$ involves an expensive matrix inversion, which makes direct computation of $\xv^{t+1}$ impractical. An inversion-free approach proposed by \cite{wei2017inner} addresses this problem by learning a convolutional neural network to approximate the matrix inversion. Note that training this neural network is data-independent since $\Umat^{-1}$ is only dependent on $\Phimat$. Applying the Sherman-Morrison-Woodbury formula, we reduce this matrix inversion to a smaller scale,
\begin{align}\label{eq:cnn}
\Umat^{-1}=\beta^{-1}(\Imat-\Phimat^H\Vmat^{-1}\Phimat),
\end{align}
where $\Vmat^{-1}=(\beta\Imat+\Phimat\Phimat^H)^{-1}$ has the dimension $m\times m$. 
To approximate $\Vmat^{-1}$, a trainable deep convolutional neural network $\Ccal_\theta$ parameterized by $\theta$ is employed, \ie, $\Ccal_\theta\approx\Vmat^{-1}$. $\Ccal_\theta$ is learned by minimizing the sum of two reconstruction losses of two auto-encoders with shared weights
\begin{align}\label{eq:train}
\arg\min_{\theta}\mathbb{E}_\epsilon[\|\epsilon-\Ccal_\theta\Vmat\epsilon\|_2^2+\|\epsilon-\Vmat\Ccal_\theta\epsilon\|],
\end{align}
where $\Vmat=\beta\Imat+\Phimat\Phimat^H$ can be computed directly, and $\epsilon$ is sampled from publicly available image datasets~\cite{imagenet_cvpr09}. Note that $\Ccal_{\theta}$ is pretrained for different $\Phimat$ and $\beta$. By plugging the learned $\Ccal_\theta$ into (\ref{eq:cnn}), we obtain a reusable term $\Umat^{-1}=\beta^{-1}(\Imat-\Phimat^H\Ccal_\theta\Phimat)$ as the replacement for the cumbersome inversion matrix. Hence $\xv^{t+1}$ is updated by
\begin{align}\label{eq:deep}
\xv^{t+1}=\beta^{-1}(\Imat-\Phimat^H\Ccal_\theta\Phimat)(\Phimat^H\yv+\beta\zv^{t+1}-\frac{\muv^{t}}{2}).
\end{align}

\subsubsection{Fourier space solution}
Equation (\ref{eq:update_x}) can be solved by transforming the problem from the image space into the Fourier space when $\Phimat$ is a partial Fourier transform matrix~\cite{Dong14_NLRCS}. For a down-sampling matrix $\Dmat$ and a Fourier transform matrix $\Fmat$, $\Phimat=\Dmat\Fmat$ is substituted into (\ref{eq:update_x})
\begin{equation}\label{eq:fft}
\xv^{t+1}=(\Dmat^H\Dmat+\beta\Imat)^{-1}(\Fmat^H\Dmat^H\yv+(\beta\zv^{t+1}-\frac{\muv^{t}}{2})),
\end{equation}
where the inverse matrix $(\Dmat^H\Dmat+\beta\Imat)^{-1}$ is a diagonal matrix and thus can be computed easily. Equation (\ref{eq:fft}) is equivalent to
\begin{equation}\label{eq:fft2}
\xv^{t+1}=\Fmat^H\{(\Dmat^H\Dmat+\beta\Imat)^{-1}(\Dmat^H\yv+\Fmat(\beta\zv^{t+1}-\frac{\muv^{t}}{2}))\}.
\end{equation}
Therefore, $\xv^{t+1}$ can be obtained by applying inverse Fourier transform to terms in the brackets of the right hand side of (\ref{eq:fft2}).

\section{Experiments \label{Sec:Results}}
\begin{table*}[t!]
	\centering
	\small
		\caption{Reconstruction PSNR (dB) and SSIM of $8$ images using different CS recovery methods at different CSr.}
	\begin{tabular}{|c|c|c|c|c|c|c|}
		\hline
		\multirow{2}{*}{Image} & \multirow{2}{*}{Method} & \multicolumn{5}{|c|}{CSr}\\
		\cline{3-7}
		& & 0.02 & 0.04 & 0.06 & 0.08 & 0.10 \\ 
		\hline
		\multirow{5}{*}{Barbara} 
		& BM3D-CS & 16.44, 0.389 & 18.03, 0.422 & 19.70, 0.470 & 21.85, 0.561 & 23.70, 0.635 \\
		& TVAL3 & 21.32, 0.588 & 22.58, 0.620 & 23.28, 0.639 & 23.78, 0.654 & 24.25, 0.667 \\
		& NLR-CS & 16.79, 0.400 & 18.90, 0.433 & 20.56, 0.501 & 24.00, 0.659 & 25.13, 0.725 \\
		& D-AMP & 15.08, 0.354 & 17.76, 0.417 & 20.44, 0.496 & 22.66, 0.623 & 24.05, 0.660 \\
		& Ours  & \textbf{26.15}, \textbf{0.812} & \textbf{27.29}, \textbf{0.835} & \textbf{27.40}, \textbf{0.837} & \textbf{28.05}, \textbf{0.848} & \textbf{28.30}, \textbf{0.852} \\
		\hline
		\multirow{5}{*}{Boats} 
		& BM3D-CS & 16.76, 0.398 & 17.98, 0.440 & 20.11, 0.507 & 23.21, 0.624 & 24.65, 0.687 \\
		& TVAL3 & 21.58, 0.634 & 23.66, 0.681 & 24.57, 0.699 & 25.22, 0.713 & 25.75, 0.725 \\
		& NLR-CS & 17.20, 0.409 & 19.41, 0.489 & 21.76, 0.559 & 23.76, 0.638 & 25.52, 0.711 \\
		& D-AMP & 15.52, 0.365 & 18.41, 0.456 & 20.76, 0.528 & 22.85, 0.600 & 25.22, 0.703 \\
		& Ours & \textbf{30.79}, \textbf{0.898} & \textbf{31.73}, \textbf{0.909} & \textbf{31.84}, \textbf{0.910} & \textbf{32.35}, \textbf{0.916} & \textbf{32.33}, \textbf{0.912} \\
		\hline
		\multirow{5}{*}{Cameraman} 
		& BM3D-CS & 16.84, 0.432 & 18.34, 0.496 & 20.62, 0.585 & 22.32, 0.654 & 23.85, 0.711 \\
		& TVAL3 & 20.33, 0.557 & 21.53, 0.579 & 22.02, 0.580 & 22.37, 0.578 & 23.19, 0.688 \\
		& NLR-CS & 17.00, 0.436 & 19.10, 0.531 & 21.60, 0.620 & 23.99, 0.709 & 26.25, 0.789 \\
		& D-AMP & 16.13, 0.401 & 17.51, 0.461 & 19.84, 0.562 & 21.74, 0.629 & 23.61, 0.705 \\
		& Ours & \textbf{25.88}, \textbf{0.828} & \textbf{26.47}, \textbf{0.829} & \textbf{26.66}, \textbf{0.823} & \textbf{27.16}, \textbf{0.824} & \textbf{27.64}, \textbf{0.819} \\
		\hline
		\multirow{5}{*}{Foreman} 
		& BM3D-CS & 18.96, 0.610 & 21.06, 0.680 & 25.48, 0.726 & 29.80, 0.831 & 31.69, 0.866 \\
		& TVAL3 & 23.35, 0.777 & 26.81, 0.808 & 28.47, 0.821 & 29.45, 0.828 & 30.02, 0.833 \\
		& NLR-CS & 19.64, 0.645 & 22.33, 0.695 & 28.09, 0.809 & 31.68, 0.865 & 34.15, 0.900 \\
		& D-AMP & 19.42, 0.639 & 22.07, 0.687 & 26.31, 0.750 & 29.12, 0.813 & 31.90, 0.872 \\
		& Ours & \textbf{34.01}, \textbf{0.934} & \textbf{34.61}, \textbf{0.936} & \textbf{35.18}, \textbf{0.937} & \textbf{35.33}, \textbf{0.939} & \textbf{36.02}, \textbf{0.939} \\
		\hline
		\multirow{5}{*}{House} 
		& BM3D-CS & 18.12, 0.518 & 20.35, 0.590 & 25.79, 0.755 & 29.78, 0.816 & 30.50, 0.824 \\
		& TVAL3 & 22.57, 0.706 & 24.58, 0.724 & 26.36, 0.740 & 27.63, 0.766 & 28.68, 0.814 \\
		& NLR-CS & 18.80, 0.554 & 21.77, 0.620 & 27.16, 0.761 & 31.23, 0.835 & 33.66, 0.869 \\
		& D-AMP & 17.74, 0.492 & 19.93, 0.577 & 24.00, 0.693 & 27.39, 0.758 & 29.84, 0.818 \\
		& Ours & \textbf{30.42}, \textbf{0.873} & \textbf{31.76}, \textbf{0.884} & \textbf{33.07}, \textbf{0.898} & \textbf{34.32}, \textbf{0.910} & \textbf{34.80}, \textbf{0.913} \\
		\hline
		\multirow{5}{*}{Lena} 
		& BM3D-CS & 16.09, 0.427 & 18.22, 0.511 & 20.47, 0.603 & 23.55, 0.694 & 24.63, 0.729 \\
		& TVAL3 & 21.60, 0.671 & 23.23, 0.693 & 23.88, 0.697 & 24.41, 0.701 & 24.81, 0.705 \\
		& NLR-CS & 16.71, 0.461 & 19.07, 0.545 & 21.36, 0.649 & 25.47, 0.746 & 25.82, 0.771 \\
		& D-AMP & 15.53, 0.399 & 17.21, 0.485 & 19.93, 0.567 & 22.29, 0.681 & 24.60, 0.733 \\
		& Ours & \textbf{28.51}, \textbf{0.881} & \textbf{29.70}, \textbf{0.894} & \textbf{29.85}, \textbf{0.889} & \textbf{30.42}, \textbf{0.898} & \textbf{30.65}, \textbf{0.896} \\
		\hline
		\multirow{5}{*}{Monarch} 
		& BM3D-CS & 14.55, 0.352 & 15.24, 0.390 & 17.65, 0.500 & 19.75, 0.581 & 21.64, 0.666 \\
		& TVAL3 & 17.31, 0.537 & 18.91, 0.577 & 19.82, 0.595 & 20.56, 0.611 & 21.24, 0.624 \\
		& NLR-CS & 14.69, 0.357 & 16.03, 0.413 & 18.53, 0.552 & 21.84, 0.686 & 25.12, 0.796 \\
		& D-AMP & 14.01, 0.334 & 14.86, 0.366 & 17.55, 0.497 & 19.84, 0.587 & 21.77, 0.675 \\
		& Ours & \textbf{26.14}, \textbf{0.884} & \textbf{27.28}, \textbf{0.895} & \textbf{28.21}, \textbf{0.901} & \textbf{28.97}, \textbf{0.904} & \textbf{29.25}, \textbf{0.904} \\
		\hline
		\multirow{5}{*}{Parrots} 
		& BM3D-CS & 16.62, 0.506 & 19.32, 0.613 & 23.07 0.693 & 25.04, 0.761 & 26.10, 0.787 \\
		& TVAL3 & 21.11, 0.714 & 22.90, 0.722 & 23.47, 0.714 & 24.38, 0.770 & 25.84, 0.804 \\
		& NLR-CS & 17.39, 0.550 & 20.84, 0.681 & 24.95, 0.782 & 28.18, 0.835 & \textbf{30.35}, 0.875 \\
		& D-AMP & 16.89, 0.520 & 19.55, 0.619 & 22.62, 0.684 & 24.10, 0.711 & 26.31, 0.792 \\
		& Ours & \textbf{28.28}, \textbf{0.899} & \textbf{28.88}, \textbf{0.893} & \textbf{29.22}, \textbf{0.890} & \textbf{29.55}, \textbf{0.887} & 29.84, \textbf{0.885} \\
		\hline
		\multirow{5}{*}{Average} 
		& BM3D-CS & 16.80, 0.454 & 18.57, 0.518 & 21.61, 0.605 & 24.42, 0.690 & 25.85, 0.738 \\
		& TVAL3 & 21.15, 0.648 & 23.02, 0.676 & 23.98, 0.685 & 24.72, 0.703 & 25.47, 0.732 \\
		& NLR-CS & 17.28, 0.477 & 19.68, 0.551 & 23.00, 0.654 & 26.27, 0.747 & 28.25, 0.805 \\
		& D-AMP & 16.29, 0.438 & 18.41, 0.509 & 21.43, 0.610 & 23.75, 0.675 & 25.91, 0.745 \\
		& Ours & \textbf{28.77}, \textbf{0.876} & \textbf{29.71}, \textbf{0.884} & \textbf{30.18}, \textbf{0.886} & \textbf{30.77}, \textbf{0.891} & \textbf{31.10}, \textbf{0.890} \\
		\hline
	\end{tabular}
	\label{tab:ours}
\end{table*}

We conduct experiments on both noisy and noiseless CS mearurements of $8$ test images: {\em Barbara}, {\em Boats}, {\em Cameraman}, {\em Foreman}, {\em House}, {\em Lena}, {\em Monarch}, and {\em Parrots}. All images are resized to $256\times 256$. 
Since excellent results have been obtained when the CS rate is large, \ie${\rm CSr}>0.1$~\cite{Dong14_NLRCS,Metzler16DAMP}, we here focus on testing cases with limited number of measurements (${\rm CSr}<0.1$). 
The CS measurements are generated by pseudo-radial sampling of the test images in the Fourier domain. Unlike traditional random sampling schemes, pseudo-radial sampling produces streaking artifacts which are more difficult to remove. 
We also perform experiments using an $M\times N$ random sensing matrix generated by a standard Gaussian distribution. The deep convolutional architecture $C_\theta$ is pretrained by the sensing matrix then used in ADMM to solve the expensive matrix inversion.
A DCT based CS recovery algorithm~\cite{Yuan16SJ} is used for the initial image estimate. 
The main parameters of the proposed method are set as follows: patch size $4\times 4$, number of similar patches for each reference patch $50$, and the imposed low rank is $20$. The regularization parameter $\eta$ and the ADMM parameter $\beta$ are tuned separately for each sensing rate. 
In practice, we have found that ADMM converges fast after a few iterations and the performance gain is mainly from our low-rank TFA. Thus we set the number of outer-loop iterations to $I=50$ and the number of inner loop iterations $J=2$. Experimental results for noiseless CS measurements and noisy CS measurements are evaluated by peak signal-to-noise ratio (PSNR) and structural similarity (SSIM)~\cite{Wang04SSIM}. 
\begin{figure}[htbp!]
	\centering
	\subfloat[]{\includegraphics[scale = 0.41]{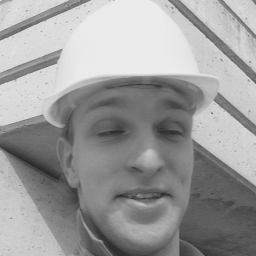}}\hspace{0.03in}
	\subfloat[]{\includegraphics[scale = 0.41]{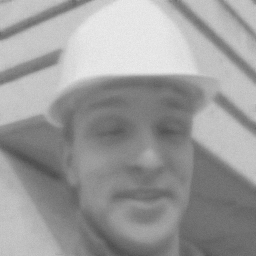}}\\
	\subfloat[]{\includegraphics[scale = 0.41]{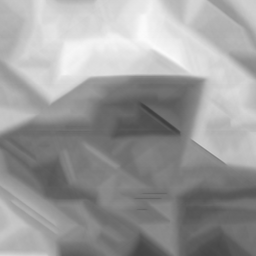}}\hspace{0.03in}
	\subfloat[]{\includegraphics[scale = 0.41]{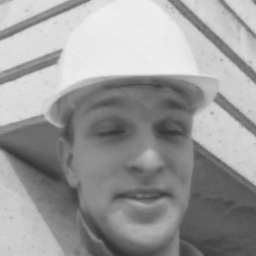}}
	\caption{Reconstruction results from noiseless CS measurements for \textit{Foreman} at CSr$=0.02$. (a) Original image; (b) TVAL3 (23.35 dB); (c) NLR-CS (19.64 dB); (d) Ours (34.01 dB).}
	\label{fig:foreman}
\end{figure}
\begin{figure}[htbp!]
	\centering
	\subfloat[]{\includegraphics[scale = 0.41]{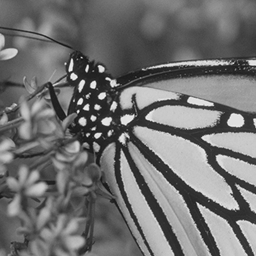}}\hspace{0.03in}
	\subfloat[]{\includegraphics[scale = 0.41]{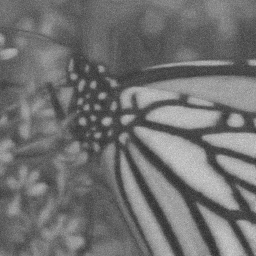}}\\
	\subfloat[]{\includegraphics[scale = 0.41]{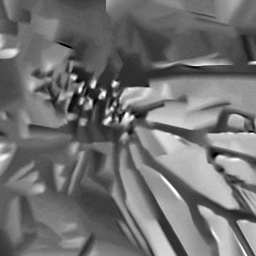}}\hspace{0.03in}
	\subfloat[]{\includegraphics[scale = 0.41]{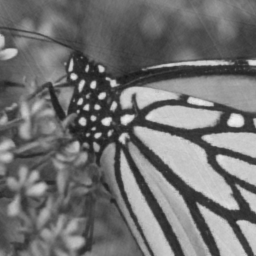}}
	\caption{Reconstruction results from noiseless CS measurements for \textit{Monarch} at CSr$=0.06$. (a) Original image; (b) TVAL3 (19.82 dB); (c) NLR-CS (18.51 dB); (d) Ours (28.21 dB).}
	\label{fig:Monarch}
\end{figure}
\subsection{Baselines}
We compare our results with four competitive CS image restoration algorithms: BM3D-CS~\cite{egiazarian2007compressed}, TVAL3~\cite{li2013efficient}, NLR-CS~\cite{Dong14_NLRCS} and D-AMP~\cite{Metzler16DAMP}. BM3D-CS is a nonparametric method that applies block matching and 3D filtering algorithm~\cite{Dabov07BM3D} on compressive sensing. TVAL3 restores images by combining the classic augmented Lagrangian multiplier method with total-variation regularization. NLR-CS is currently a state-of-the-art method that uses a nonlocal low-rank regularization method along with ADMM to solve image CS problems. D-AMP employs a denoiser in an approximate message passing framework, achieving state-of-the-art performance on noisy measurements, especially at high CS rates. The source codes of these baseline methods are downloaded from the respective author's website and parameters for these algorithms are set to their default values.

\subsection{Experiments with Noiseless CS Measurements}
We first perform CS image restoration experiments from noiseless measurements using pseudo-radial sampling scheme at five different CS rates, \ie, CSr = \{$0.02$, $0.04$, $0.06$, $0.08$, $0.10$\}. Table~\ref{tab:ours} summarizes the results of our proposed algorithm compared with various CS inversion algorithms at different CSr values. Both BM3D-CS and the state-of-the-art methods NLR-CS and D-AMP suffer at extremely low CSr. As CSr increases, NLR-CS yields significant performance improvements, which leads one of its results at higher CSr ($0.10$) to surpass other methods. Our proposed method achieves the best performance in all cases when CSr$<0.1$. On average, our proposed NLR-TFA algorithm outperforms all other competing methods at CSr = \{$0.02$, $0.04$, $0.06$, $0.08$, and $0.10$\}. The PSNR gains of our NLR-TFA over BM3D-CS, TVAL3, NLR-CS and D-AMP can be as much as $15.04$ dB, $10.65$ dB, $14.36$ dB, and $14.58$ dB, respectively. Furthermore, it can be observed that our average reconstruction PSNR only decreases by $2.33$ dB as CSr decreases from $0.10$ to $0.02$, while this number for NLR-CS and D-AMP are $10.67$ dB and $9.62$ dB, respectively, indicating that the proposed method is very stable at low CSr, \ie, with limited number of measurements. 

To evaluate the reconstruction visually, two examples of restored \textit{Foreman} and \textit{Monarch} images at CSr of $0.02$ and $0.06$ are shown in Figs.~\ref{fig:foreman} and \ref{fig:Monarch}. It is evident that our method recovers the best visual quality among all competiting methods. Large-scale sharp edges and small-scale fine structures are both reconstructed in two images. In particular, at extremely low CSr of $0.02$, our NLR-TFA method (Fig.~\ref{fig:foreman}(d)) can effectively approximate the original image while some other methods, such as NLR-CS, can only reconstruct scratches (Fig.~\ref{fig:foreman}(c)).
\begin{table}[htbp!]
	\centering
	\caption{PSNR(dB) of reconstructions from measurements generated by random Gaussian sampling of 8 images at different CSr.}
	\begin{tabular}{|c|c|c|c|c|c|}
		\hline
		\multirow{2}{*}{Image} & \multirow{2}{*}{Method} & \multicolumn{4}{|c|}{CSr}\\
		\cline{3-6}
		& & 0.02 & 0.04 & 0.06 & 0.08 \\ 
		\hline
		\multirow{2}{*}{Barbara} 
		& D-AMP & 15.79 & 17.56 & 19.20 & 21.96 \\
		& Ours & \textbf{24.13} & \textbf{25.52} & \textbf{26.38} & \textbf{27.22} \\
		\hline
		\multirow{2}{*}{Boats} 
		& D-AMP & 16.06 & 17.93 & 19.84 & 22.47 \\
		& Ours & \textbf{26.04} & \textbf{27.44} & \textbf{28.59} & \textbf{29.51} \\
		\hline
		\multirow{2}{*}{Cameraman} 
		& D-AMP & 16.66 & 18.32 & 19.79 & 22.42 \\
		& Ours & \textbf{23.59}\ & \textbf{24.66}\ & \textbf{25.41} & \textbf{26.15} \\
		\hline
		\multirow{2}{*}{Foreman} 
		& D-AMP & 20.50 & 22.78 & 25.75 & 28.40 \\
		& Ours  & \textbf{27.72} & \textbf{29.23} & \textbf{30.66} & \textbf{31.47} \\
		\hline
		\multirow{2}{*}{House} 
		& D-AMP & 18.61 & 20.85 & 24.42 & 26.97 \\
		& Ours & \textbf{26.68} & \textbf{28.20} & \textbf{29.48} & \textbf{30.17} \\
		\hline
		\multirow{2}{*}{Lena} 
		& D-AMP & 16.34 & 18.15 & 19.77 & 22.15 \\
		& Ours & \textbf{24.63} & \textbf{26.04} & \textbf{26.99} & \textbf{27.78} \\
		\hline
		\multirow{2}{*}{Monarch} 
		& D-AMP & 14.50 & 15.68 & 17.94 & 19.93 \\
		& Ours & \textbf{21.57} & \textbf{23.01} & \textbf{24.19} & \textbf{24.89} \\
		\hline
		\multirow{2}{*}{Parrots} 
		& D-AMP & 17.49 & 19.38 & 22.56 & 25.24 \\
		& Ours & \textbf{25.92} & \textbf{27.19} & \textbf{27.96} & \textbf{28.54} \\
		\hline
		\multirow{2}{*}{Average} 
		& D-AMP & 16.99 & 18.83 & 21.15 & 23.69 \\
		& Ours & \textbf{25.04} & \textbf{26.41} & \textbf{27.46} & \textbf{28.22} \\
		\hline
	\end{tabular}
	\label{tab:deep}
\end{table}

We next generate CS measurements by sampling the test images with a random Gaussian sensing matrix. A deep convolutional neural network $\Ccal_\theta$ is pretrained using the sensing matrix to solve the large-scale matrix inversion problem in ADMM. We compare our results with D-AMP which provides random Gaussian sensing implementations in their sourcing code. The reconstruction results at CSr of $0.02$, $0.04$, $0.06$ and $0.08$ are shown in Table~\ref{tab:deep}. Our proposed method outperforms D-AMP on all test images and sensing rates. In addition, the runtime of the algorithm using this inversion-free approach is also much faster than using approaches with inner-loop updates, such as conjugate gradient (CG). This demonstrates the efficiency and accuracy of the deep convolutional architecture on approximating the inversion matrix.
\begin{figure*}[htbp!]
	\centering
	\subfloat[]{\includegraphics[scale = 0.37]{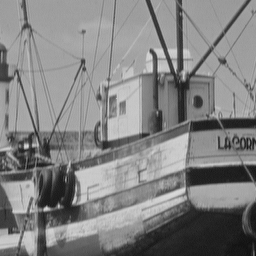}}\hspace{0.02in}
	\subfloat[]{\includegraphics[scale = 0.37]{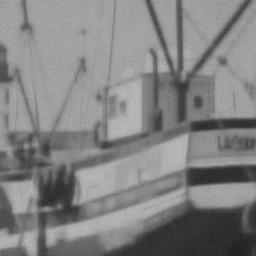}}\hspace{0.02in}
	\subfloat[]{\includegraphics[scale = 0.37]{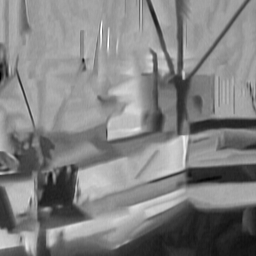}}\hspace{0.02in}
	\subfloat[]{\includegraphics[scale = 0.37]{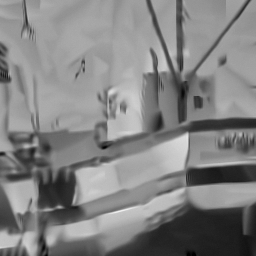}}\hspace{0.02in}
	\subfloat[]{\includegraphics[scale = 0.37]{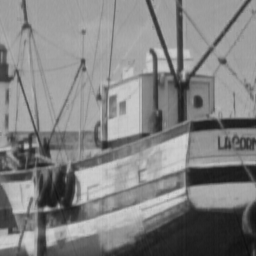}}
	\caption{Reconstruction results of \textit{Boats} images from noisy CS measurements with noise standard deviation equal to $30$ at CSr$=0.10$. (a) Original image; (b) TVAL3 (24.02 dB); (c) NLR-CS (23.48 dB); (d) DAMP (24.71 dB); (e) Ours (30.97 dB).}
	\label{fig:boats}
\end{figure*}
\begin{figure}[!htbp]
	\centering
	\subfloat[]{\includegraphics[scale = 0.4]{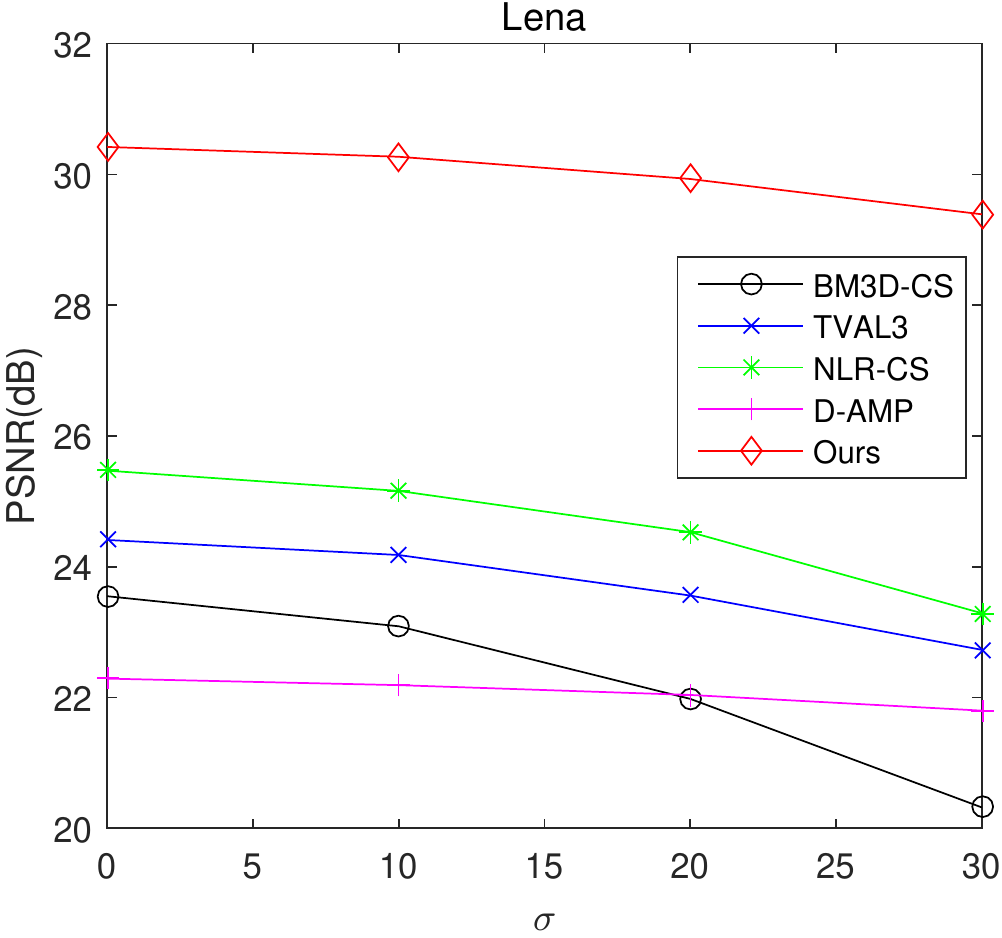}}\hspace{0.03in}
	\subfloat[]{\includegraphics[scale = 0.4]{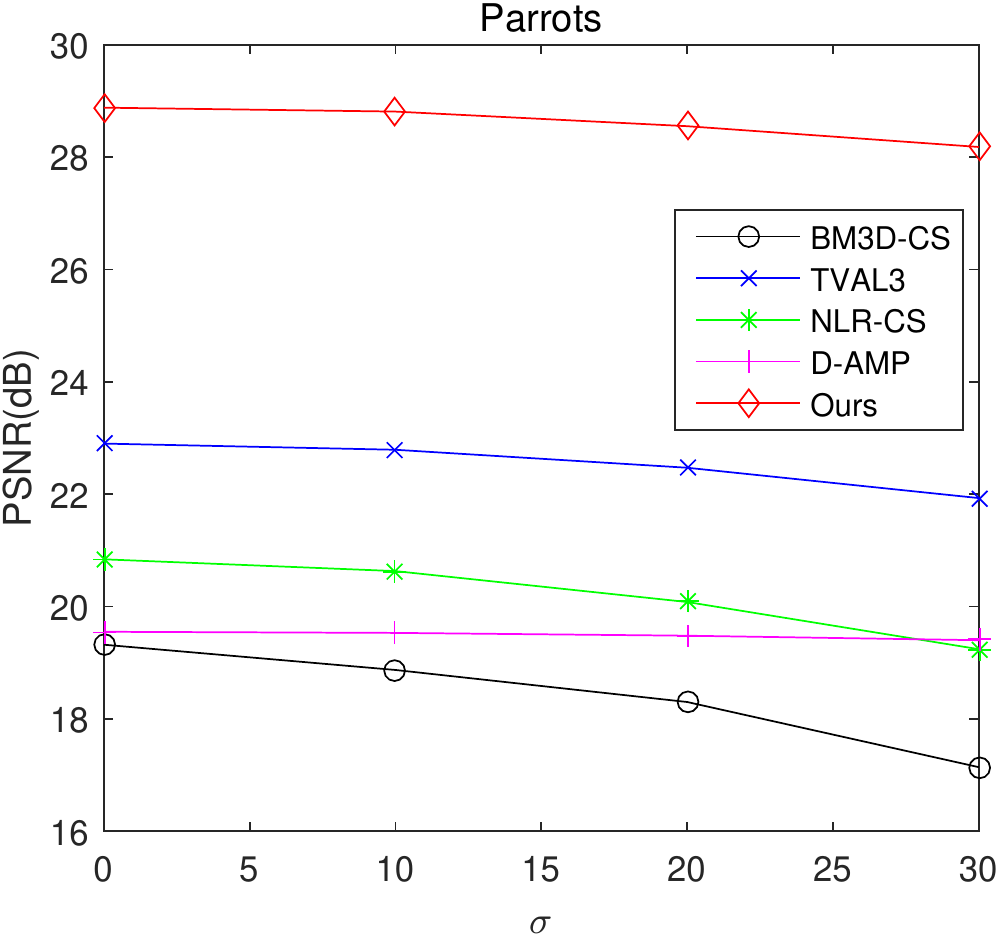}}
	\caption{Comparison of different CS methods with noisy measurements for \textit{Lena} and \textit{Parrots} images. (a) CSr$=0.08$. (b) CSr=$0.04$.}
	\label{fig:noise}
\end{figure}
\begin{table}[htbp!]
	\centering
	\caption{PSNR(dB) of reconstructions with variant patch sizes at different CSr for \textit{foreman} image.}
	\begin{tabular}{|c|c|c|c|c|c|}
		\hline
		\multirow{2}{*}{Patch Size} & \multicolumn{5}{|c|}{CSr}\\
		\cline{2-6}
		& 0.02 & 0.04 & 0.06 & 0.08 & 0.10 \\ 
		\hline
		$4\times4$ & 34.01 & 34.61 & 35.18 & 35.33 & 36.02 \\
		\hline
		$6\times6$ & 31.16 & 31.95 & 32.72 & 32.88 & 33.59 \\
		\hline
		$8\times8$ & 29.63 & 30.41 & 31.37 & 31.68 & 32.24 \\
		\hline
		$10\times10$ & 28.57 & 29.45 & 30.44 & 30.86 & 31.33 \\
		\hline
		$12\times12$ & 27.39 & 28.61 & 29.60 & 30.11 & 30.54 \\
		\hline
	\end{tabular}
	\label{tab:size}
\end{table}

\paragraph{Patch Size Selection}
We have conducted experiments using different patch sizes. As can be seen from Table \ref{tab:size} (in the pseudo-radial sampling scheme), smaller patch sizes lead to better reconstruction results. When we adopt patch size of $4\times4$ on the \textit{foreman} image, the PSNR is significantly increased compared to results with larger patch sizes. Similar conclusions are found on other images as well.
In addition, the running time of our model is largely influenced by the tensor size. For tensor size of $4\times4\times50$ (image patch size $4\times4$ and $50$ similar patches for each reference patch), the average running time at CSr=$0.06$ is about 10 minutes on a desktop with i7 CPU @3.4GHz and 24G RAM.

\subsection{Experiments with Noisy CS Measurements}
Similar experiments are conducted with noisy CS measurements to show the robustness of our algorithm to noise. The noisy CS measurements are obtained by adding random zero-mean Gaussian noise to the noiseless CS measurements. We test the algorithms at three levels of noise, with standard deviation $\sigma$=\{$10$, $20$, $30$\}. The PSNR comparison of all methods for \textit{Boats} and \textit{Monarch} images at sensing rates of $0.10$ and $0.04$, respectively, are shown in Fig.~\ref{fig:noise}. Our proposed NLR-TFA outperforms other competing methods at all levels of noise. Furthermore, D-AMP shows great robustness to noise while BM3D-CS and NLR-CS are relatively sensitive to high-level noise. As the noise level increases, the reconstruction performance of our algorithm decreases slowly. This shows that the proposed method is robust to noise. In Fig.~\ref{fig:boats}, we show the reconstructions of \textit{Boats} images using various algorithms at CSr=$0.10$ and $\sigma=30$. The proposed NLR-TFA clearly yields the best reconstruction and is robust to noise. 
%

\section{Conclusion \label{Sec:Con}}
We have presented a low-rank tensor-factor-analysis-based approach to solve image-restoration problems. A tensor is generated by concatenating similar patches from the estimated image for each exemplar patch. Low-rank is imposed on the tensors to exploit non-local correlation and the high-order structure information via tensor factorization. ADMM is employed on low-rank tensors, where we use either a pretrained convolutional architecture or Fourier approaches to solve a matrix inversion. Experimental results demonstrate that the proposed NLR-TFA method outperforms state-of-the-art algorithms on CS image reconstruction at low CS sampling rates.

We have demonstrated the superiority of our model in image CS with three-way tensors, which can also be used in depth CS~\cite{Llull15Optica,Llull14COSI,Yuan16AO}, polarization CS ~\cite{Tsai15OE} and dynamic range CS~\cite{Yuan_16_OE}. 
Future work includes extending the model to four-way tensors for video CS~\cite{Patrick13OE,Llull2015_book,Sun16OE,Sun17OE,Yang14GMMonline,Yang14GMM,Yuan14CVPR,Yuan&Pang16_ICIP,Yuan16BOE,Yuan16COSI,Yuan17_COSI_rgbD}, hyperspectral image CS~\cite{Cao16SPM,Yuan15JSTSP} and to five-way tensors for joint spectral-temporal CS~\cite{Tsai15COSI,Tsai15OL}.

{\small
\bibliographystyle{ieee}
\bibliography{LR-TFA}
}

\end{document}